# Learning Random Numbers to Realize Appendable Memory System for Artificial Intelligence to Acquire New Knowledge after Deployment

Kazunori D Yamada*

*Graduate School of Information Sciences, Tohoku University, Sendai, Japan*
*Unprecedented-scale Data Analytics Center, Tohoku University, Sendai, Japan*

In this study, we developed a learning method for constructing a neural network system capable of memorizing data and recalling it without parameter updates. The system we built using this method is called the Appendable Memory system. The Appendable Memory system enables an artificial intelligence (AI) to acquire new knowledge even after deployment. It consists of two AIs: the Memorizer and the Recaller. This system is a key–value store built using neural networks. The Memorizer receives data and stores it in the Appendable Memory vector, which is dynamically updated when the AI acquires new knowledge. Meanwhile, the Recaller retrieves information from the Appendable Memory vector. What we want to teach AI in this study are the operations of memorizing and recalling information. However, traditional machine learning methods make AI learn features inherent in the learning dataset. We demonstrate that the systems we intend to create cannot be realized by current machine learning methods, that is, by merely repeating the input and output learning sequences with AI. Instead, we propose a method to teach AI to learn operations, by completely removing the features contained in the learning dataset. Specifically, we probabilized all the data involved in learning. This measure prevented AI from learning the features of the data. The learning method proposed in the study differs from traditional machine learning methods and provides fundamental approaches for building an AI system that can store information in a finite memory and recall it at a later date.

KEYWORDS:  learning with random values, connectionism, Appendable Memory, key–value store, in-context learning

## 1 Introduction

In this study, we propose a novel learning method to teach an artificial intelligence (AI) to learn how to operate. Specifically, we trained the AI to memorize information and to recall that information. The operations performed by the AI developed with our learning method are independent of the property of the data used for training, allowing the AI to memorize new information and utilize it even after deployment.

The development of an AI, which continues to learn even after deployment, is a challenging problem in the field of AI research [1]. AI that can continue to learn post deployment must possess the following abilities:

(1) Memorizing: the ability to retain information that has been supplied thus far in memory data and to add the newly supplied information to existing memory data without compromising the already-memorized information,
(2) Recalling: the ability to freely extract necessary information from memory data.

In this study, the AI system we aim to develop is one that can memorize and recall information without any additional learning processes involving parameter updates after deployment. We intend to teach AI two operations: memorizing and recalling.

Specifically, the system we aim to construct in the study is shown in Figure 1. We henceforth refer to this system as the Appendable Memory system. The Appendable Memory system consists of a series of two AIs: the Memorizer and the Recaller. During the memorization phase, the Memorizer receives multiple pieces of information and generates an Appendable Memory, which encapsulates various pieces of information. The Appendable Memory allows for the addition of new information. During the recall phase, the Recaller uses this memory vector and a piece of data that triggers the system to restore information from the memory vector. The Memorizer and the Recaller are already trained AIs that do not update their parameters when memorizing or recalling information. Instead, the Appendable Memory changes each time the Memorizer adds new information to it.

Next, we consider actually training this system, but this cannot be achieved with current machine learning methods. This is because machine learning merely teaches AI to elucidate features inherent in learning datasets, not to learn operations. In the memorization phase of Figure 1, the input vector is composed of keys and values. Imagine a scenario where these keys and values have no predictable pattern, meaning they are random. In such a case, if the Recaller is trained repeatedly, it might improve prediction performance only for the training dataset. This is because the Recaller only needs to remember all combinations of keys and values in the dataset. However, if a completely new key is input, the Recaller can only return a random value, based on the premise that there is no rule in the combination of keys and values. We demonstrate this by an experiment. Through the experiment, we demonstrate that, while current machine learning methods can extract patterns inherent in data, they cannot teach AI to learn operations, that is, how to remember or recall information.

---

*Corresponding author. E-mail: yamada@tohoku.ac.jp

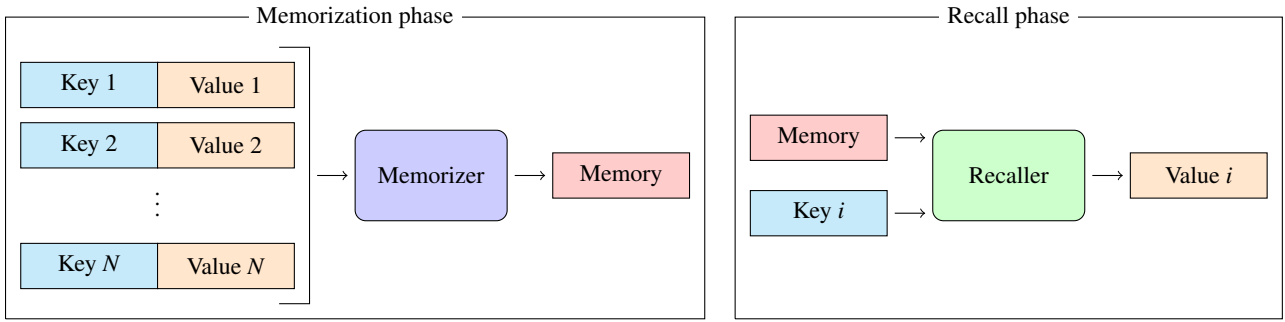

Fig. 1: Details of computation in the Appendable Memory system. In the memorization phase shown in the left panel, $N$ pairs of key and value combinations are used as inputs to the Memorizer in no particular order, which results in an output value, which is the Appendable Memory vector. In the recall phase shown in the right panel, the memory vector and a key, which are parts of the input information from the memorization process, are used as inputs to the Recaller to output the value vector corresponding to the key vector. In the diagram, rectangles represent data, and rounded rectangles represent operations.

Originally, machine learning methods fit AI to features inherent in a dataset. However, the system we intend to build cannot be realized by the usual machine learning methods. We want the Appendable Memory system not to learn features in the dataset. Therefore, to delete predictable patterns in the dataset, we probabilized the data contained in the dataset. By generating all data randomly at each learning step, namely epoch, we prevented AI from learning features in the training dataset. As a result, we succeeded in designing a system that can encode multiple pieces of information independently into a memory vector and restore the information freely.

In this article, we will first discuss the limitations of current machine learning methods, then propose a machine learning method using random numbers to realize a memory to which new information can be appended, and finally describe applications that utilize Appendable Memory to implement a sorting algorithm.

## 2 Related Work

The method we propose in this study involves probabilizing the learning data during neural network training. Typically, supervised machine learning methods teach AI to learn the features of data in the training dataset. Such AI makes predictions based on the features of the training dataset. In contrast, the AI we intend to build is one capable of memorizing and recalling information, regardless of any properties in the training dataset. Therefore, to teach AI to only learn the operations, we probabilized the dataset for each parameter update during learning, to remove its inherent features.

There are several methods that apply stochastic concepts for AI learning. The most basic is using random numbers for parameter initialization. A survey paper [2] introduced various methods that utilize randomness. However, there is no research and development on probabilizing the dataset itself.

Usually, stochastic gradient descent (SGD) is used in neural network training [3]. SGD involves randomly shuffling and batching the training dataset. This is not the same as randomly generating the training dataset itself, as we do in this study. SGD only shuffles the training dataset and samples portions of the dataset for each parameter update randomly.

For other types of neural networks such as generative adversarial networks (GANs) [4] and diffusion models [5], the concept of randomness is used for the learning process. However, these differ from our method. Generative models like GANs and diffusion models utilize random noise to ensure the diversity of the generated data. In contrast, our method uses randomness not to teach the AI the features inherent in the dataset, but to teach it operations. Reinforcement learning can be used to make models acquire operations instead of learning patterns in the data, and therefore the underlying mechanism of the method we developed might be similar to that of reinforcement learning in this regard.

## 3 Experiments

### 3.1 Network Architecture

This study proposes an AI system for storing information within a single, modality-unrestricted finite-sized vector and retrieving information from it by utilizing a neural network resembling a recurrent neural network (RNN) and an encoder–decoder network, referred to as Memorizer–Recaller. The proposed system can generate memory data, an Appendable

Memory vector, which is updated dynamically when the AI acquires new knowledge. The learning method in this study differs from traditional machine learning methods and provides fundamental approaches for building an AI system that can store information within finite medium and can freely utilize them in the future.

The aim of this study is to train the Memorizer–Recaller network such that the Memorizer can store $N$ pairs of vectors into a memory vector, while the Recaller outputs the corresponding value from the memory vector along with one key vector, as shown in Figure 1.

Here, we presented the formulas of the models. The left arrow ($\leftarrow$) is a symbol that assigns the value on its right to the variable on its left. The first model, that is, the Memorizer, is represented by the following formula:

$$u_t \leftarrow f(k_t, v_t), \tag{3.1}$$

$$p \leftarrow \sigma(w_1 u_t + b_1), \tag{3.2}$$

$$q \leftarrow \sigma(w_2 m_{t-1} + b_2), \tag{3.3}$$

$$m_t \leftarrow \sigma(w_3(p+q) + b_3), \tag{3.4}$$

where $w$ denotes the weight parameter, $b$ is the bias parameter, and $\sigma$ is the activation function. We used LeakyReLU [6] as the activation function. In addition, $f$ is a function that concatenates two vectors, $t$ is an integer from 1 to $N$, $k_t$ is the $t$-th key, $v_t$ is the $t$-th value, and $m_t$ is the $t$-th memory vector. In the Memorizer, the $(t-1)$-th memory vector is used to generate the $t$-th memory vector, and the Memorizer has a recurrent loop in its structure; therefore, the Memorizer is a type of RNN. The variable $p$ in the final formula is derived from the input data, whereas $q$ is derived from the $(t-1)$-th memory vector. The memory vector includes both previous memory information and new input information. Because calculating the elementwise summation of $p$ and $q$ is essential, these vectors must have identical dimensions. We perform computations for all $N$ data points during the memorization phase to obtain $m_N$. The second model, that is, the Recaller, is represented by the following formula:

$$r \leftarrow \sigma(w_4 k_i + b_4), \tag{3.5}$$

$$s \leftarrow \sigma(w_5 m_N + b_5), \tag{3.6}$$

$$\hat{v}_i \leftarrow f(r, s), \tag{3.7}$$

$$\hat{v}_i \leftarrow \sigma(w_6 \hat{v}_i + b_6), \tag{3.8}$$

$$\hat{v}_i \leftarrow \phi(w_7 \hat{v}_i + b_7), \tag{3.9}$$

where $\phi$ is the activation function, and we used the softmax function, because in the experiment the Recaller is used to classify and predict values from 0 to 9. In addition, $k_i$ is any one of the $N$ keys used as the input to the Memorizer. In the recall phase, it attempts to output $\hat{v}_i$ corresponding to $k_i$.

Figure 2 illustrates the Memorizer–Recaller network. In the figure, the structure on the left is the Memorizer, and the one on the right is the Recaller. The Memorizer accepts pairs of key and value vectors as inputs and outputs a memory vector. Repeating this calculation for all $N$ input values ultimately generates a memory vector that is expected to contain all input information. The Recaller accepts a single key and the memory vector generated by the Memorizer as input and outputs the value corresponding to the key.

The objective to train the Memorizer–Recaller network is to develop an AI system that can realize the storage and retrieval of multiple pieces of information, as shown in Figure 1. In this study, we have not discussed the performance enhancement of the system; therefore, we did not include any luxury mechanisms, which are generally used to improve the performance of neural networks. We designed the system with a simple combination of layers, as illustrated in the figure. The source codes used to generate all the networks in this study are available in a GitHub repository, https://github.com/yamada-kd/Memorizer-recaller.git.

### 3.2 Learning Process of the Models

#### 3.2.1 Learning in a Standard Manner

Firstly, we trained the model in a standard learning manner. Various hyperparameters were used to train the Appendable Memory system in the standard manner and these will be explained before proceeding to explain the training methodology. The size of the intermediate layer of both the Memorizer and the Recaller was set to 256. Similarly, the dimension of the memory vector, which is the output of the Memorizer, was set to 256. Because the Recaller is a predictor that outputs integers from 0 to 9, the dimension of its output vector was set to 10. In addition, the dimension of the key vector was set to 16. Each element of the key vector is a floating-point number arbitrarily generated from a continuous uniform distribution, followed by parameters with a minimum value of 0 and maximum value of 9. The value vector is one-dimensional, and its only element is an integer randomly generated from a discrete uniform distribution with a minimum value of 0 and a maximum value of 9. The volume of data to be input to the Memorizer is $N$; $N$ is variable for the experiment. For training, 1,024 combinations of these $N$ input data were generated and used in the batch learning method. During the learning process of the Memorizer, the value of $t$ was varied from 1 to $N$, the memory vector $m_t$ was calculated, and finally, $m_N$ was computed. For $m_0$, random numbers generated from a uniform distribution following the parameters with a minimum value of -1 and a maximum value of 1 were used. Because the softmax function was used in the output layer, the cross-entropy error function was used as the cost function. The parameters were updated when the error between the value and teacher data was calculated after inputting $m_N$ and each of the $N$ keys to the Recaller to output the predicted value. Therefore, the cost function $L$ was computed as follows:

$$m_t = M(k_t, v_t, m_{t-1}), \tag{3.10}$$

$$\hat{v}_i = R(k_i, m_N), \tag{3.11}$$

$$L(\mathbf{w}, \mathbf{b}) = \frac{1}{N} \sum_{i=1}^{N} l(\hat{v}_i, v_i), \tag{3.12}$$

where $M$ represents the Memorizer, $R$ represents the Recaller, $l$ denotes the cross-entropy error function, $\mathbf{w}$ signifies all weight parameters of the Memorizer–Recaller network, and $\mathbf{b}$ denotes all bias parameters of the Memorizer–Recaller network. The evaluation of the model was performed using accuracy. The Recaller outputs $\hat{v}_i$ for each input $k_i$, and it was considered correct if this value equaled $v_i$. This calculation was repeated 1,024 times, the sample size, and the average value was taken as the final accuracy. Since this is a 10-class classification problem, the value would be 0.1 if the Recaller made random predictions. We used the Adam [7] as the parameter optimization method, with all hyperparameters, including a learning rate of 0.001, set to the default values in TensorFlow 2.2.0.

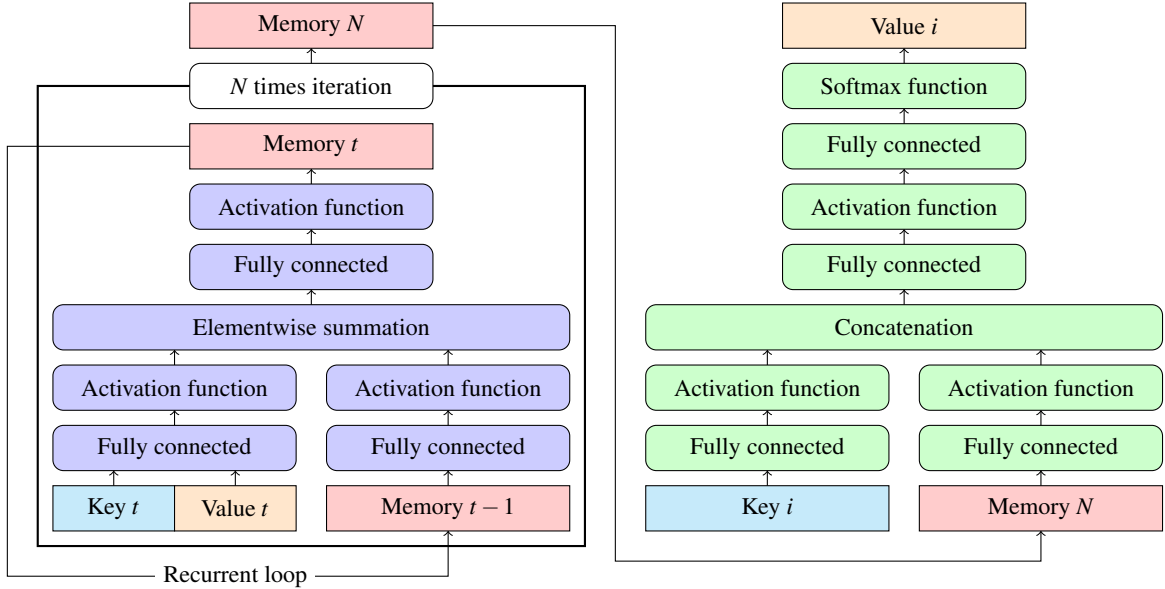

Fig. 2: Illustration of the Memorizer–Recaller network. The network on the left is the Memorizer, which accepts $N$ pairs of key and value vectors as input, and outputs the memory vector. The Recaller on the right accepts a single key and the memory vector as input and outputs the corresponding value. The rectangles represent data and rounded rectangles represent operations.

The results of the training are listed in Table 1. For the training, the training and validation datasets were randomly generated. In other words, there is no discernible rule or pattern that the machine learning model can discover in each dataset. Therefore, to derive the correct answer in the validation phase, the Recaller must refer to the memory vector generated throughout the memorization phase. The training was terminated when the accuracy in the training phase reached 0.8. In this case, we stopped the training based on the accuracy on the training dataset. The reason is that this model exhibited overfitting, where the accuracy on the validation dataset did not improve at all, while only the accuracy on the training dataset increased. Table 1 displays the number of epochs required and the accuracy of the validation dataset. Here, an epoch refers to the unit of time during which the models process all the data, that is, 1,024 sets of sampling data in this experiment. As mentioned earlier, each sampling dataset consists of $N$ pairs of keys and values. As a result of the training, the accuracy in the training phase was 0.8; however, the accuracy in the validation phase was still low; thus, we could not build a sufficiently good model.

It was impossible to construct a robust model by the training because the model overfitted the training dataset and adapted excessively to it. In other words, this model learned only the characteristics of the training dataset consisting of random elements. Therefore, it could not adapt to the validation dataset, which had no correlation and did not share any characteristics with the training dataset. Although the Recaller should ideally refer to the memory vector, which is a condensed medium of the information input to the Memorizer, to derive correct answers, it merely attempts to output values only based on the pattern of the randomly generated vector.

| N | Epoch | Training accuracy | Validation accuracy |
|---|-------|-------------------|---------------------|
| 2 | 550   | 0.802             | 0.233               |
| 3 | 650   | 0.802             | 0.168               |
| 4 | 820   | 0.800             | 0.146               |
| 5 | 1000  | 0.801             | 0.139               |
| 6 | 1100  | 0.801             | 0.120               |
| 7 | 1510  | 0.800             | 0.118               |
| 8 | 1470  | 0.801             | 0.113               |
| 9 | 1720  | 0.800             | 0.111               |

Table 1: Results of training with various numbers of input data $N$. Training was terminated when the accuracy for the training dataset reached 0.8. Epochs represent the learning time required for the accuracy of the models to reach this point.

Basically, incorporating the function of referencing to the memory vector is impossible because normally machine learning models only generate output values corresponding to the input values. The parameters of neural networks trained by the standard learning method memorize the relationship between the key and value. Once a system is overfitted to the training dataset, it need not refer to the memory vector. AI constructed with the conventional machine learning method can return appropriate output values for input values, and it does not acquire the ability of operation to store knowledge in the memory vector or extract information from it. In short, we could not develop the Appendable Memory system shown in Figure 1 by the conventional machine learning method.

### 3.2.2 Learning with Random Values

The objective of a Memorizer–Recaller network is to imitate two operations: memorizing input values to the Appendable Memory vector, and searching for necessary information from the memory and recalling it. In the conventional machine learning training method described in the previous section, a model learns the characteristics of the data and attempts to understand the patterns inherent to them. In contrast, this study aims to construct a system that can memorize and recall information. The proposed system in the study does not search for patterns inherent to the data but is a type of mnemonic system to learn how to memorize information.

In this study, we removed the patterns from the learning data to prevent the model from recognizing patterns inherent in the data and instead enable it to learn how to operate. Specifically, we probabilized $k_i$, which is the input value to the Memorizer and Recaller, and $v_i$, which is the input value to the Memorizer and teacher data for the Recaller. During the learning process, each of the 16 elements of $k_i$ were randomly generated from a continuous uniform distribution, with parameters ranging from 0 to 9 for each epoch. Only the element of the value vector was generated from a discrete uniform distribution with a minimum value of 0 and a maximum value of 9. We used randomly generated data in the learning process, and the aim was to ensure that random data do not have any patterns. The objective of the study was to train a Memorizer–Recaller network that could acquire the skills of remembering and retrieving information from memory, rather than simply learning the underlying patterns of data.

The results of learning using this method are listed in Table 2. The learning process ended when the validation accuracy reached 0.8. In general, the accuracy of the validation dataset is lower than that of the training dataset. If this is not the case, abnormal learning settings may be suspected. However, because both the training dataset and validation dataset in this learning method are random values, if the learning progresses normally in the training dataset, the accuracy in the validation dataset improves simultaneously, and this behavior is normal.

We conducted training while varying the volume of data $N$ to be memorized from 2 to 9. Until $N$ reached 8, the accuracy reached 0.8 for both the training and validation datasets. In other words, the trained Memorizer–Recaller was able to derive the correct answer in a validation dataset by learning from a training dataset where both the training and validation dataset were randomly generated and thus had no patterns. However, when $N$ was 9, despite training the model for 500,000 epochs, the accuracy did not reach 0.8, indicating that the learning process did not proceed successfully. In other words, the Memorizer–Recaller constructed under these learning conditions could memorize at most 8 pieces of information.

Next, to demonstrate the performance of the Memorizer–Recaller, we sequentially input 8 sets of key and value data into the Memorizer to generate a memory vector, as illustrated in the upper panel of Figure 3. Subsequently, we conducted a test using all keys to determine whether the values could be retrieved using the memory vector and Recaller. The

| $N$ | Epoch | Training accuracy | Validation accuracy |
|---|---|---|---|
| 2 | 5580 | 0.800 | 0.800 |
| 3 | 9200 | 0.800 | 0.800 |
| 4 | 13920 | 0.800 | 0.800 |
| 5 | 19800 | 0.800 | 0.800 |
| 6 | 30600 | 0.800 | 0.800 |
| 7 | 53140 | 0.800 | 0.800 |
| 8 | 95280 | 0.800 | 0.800 |
| 9 | 500000 | 0.259 | 0.259 |

Table 2: Results of training with input data $N$ where the training dataset is randomly generated for each epoch. Training was terminated when the accuracy for the validation dataset reached 0.8. Epochs represent the learning time required for the accuracy of the models to reach this point.

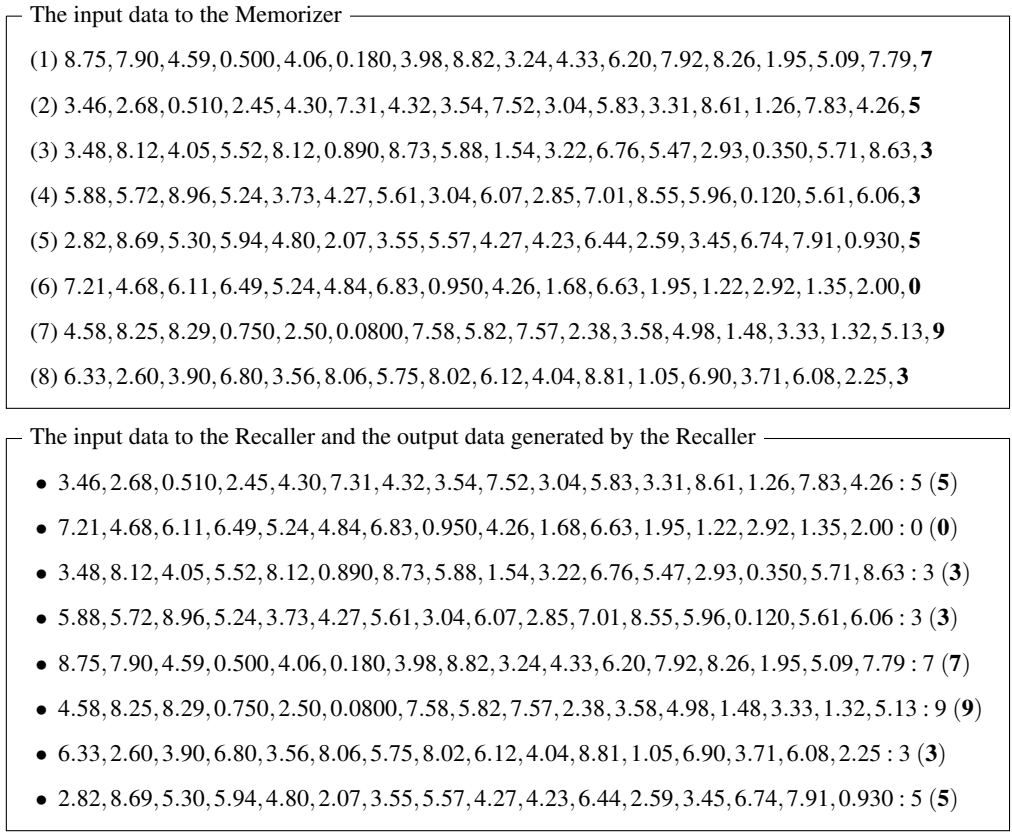

The input data to the Memorizer

(1) 8.75, 7.90, 4.59, 0.500, 4.06, 0.180, 3.98, 8.82, 3.24, 4.33, 6.20, 7.92, 8.26, 1.95, 5.09, 7.79, **7**

(2) 3.46, 2.68, 0.510, 2.45, 4.30, 7.31, 4.32, 3.54, 7.52, 3.04, 5.83, 3.31, 8.61, 1.26, 7.83, 4.26, **5**

(3) 3.48, 8.12, 4.05, 5.52, 8.12, 0.890, 8.73, 5.88, 1.54, 3.22, 6.76, 5.47, 2.93, 0.350, 5.71, 8.63, **3**

(4) 5.88, 5.72, 8.96, 5.24, 3.73, 4.27, 5.61, 3.04, 6.07, 2.85, 7.01, 8.55, 5.96, 0.120, 5.61, 6.06, **3**

(5) 2.82, 8.69, 5.30, 5.94, 4.80, 2.07, 3.55, 5.57, 4.27, 4.23, 6.44, 2.59, 3.45, 6.74, 7.91, 0.930, **5**

(6) 7.21, 4.68, 6.11, 6.49, 5.24, 4.84, 6.83, 0.950, 4.26, 1.68, 6.63, 1.95, 1.22, 2.92, 1.35, 2.00, **0**

(7) 4.58, 8.25, 8.29, 0.750, 2.50, 0.0800, 7.58, 5.82, 7.57, 2.38, 3.58, 4.98, 1.48, 3.33, 1.32, 5.13, **9**

(8) 6.33, 2.60, 3.90, 6.80, 3.56, 8.06, 5.75, 8.02, 6.12, 4.04, 8.81, 1.05, 6.90, 3.71, 6.08, 2.25, **3**

The input data to the Recaller and the output data generated by the Recaller

- 3.46, 2.68, 0.510, 2.45, 4.30, 7.31, 4.32, 3.54, 7.52, 3.04, 5.83, 3.31, 8.61, 1.26, 7.83, 4.26 : 5 (**5**)
- 7.21, 4.68, 6.11, 6.49, 5.24, 4.84, 6.83, 0.950, 4.26, 1.68, 6.63, 1.95, 1.22, 2.92, 1.35, 2.00 : 0 (**0**)
- 3.48, 8.12, 4.05, 5.52, 8.12, 0.890, 8.73, 5.88, 1.54, 3.22, 6.76, 5.47, 2.93, 0.350, 5.71, 8.63 : 3 (**3**)
- 5.88, 5.72, 8.96, 5.24, 3.73, 4.27, 5.61, 3.04, 6.07, 2.85, 7.01, 8.55, 5.96, 0.120, 5.61, 6.06 : 3 (**3**)
- 8.75, 7.90, 4.59, 0.500, 4.06, 0.180, 3.98, 8.82, 3.24, 4.33, 6.20, 7.92, 8.26, 1.95, 5.09, 7.79 : 7 (**7**)
- 4.58, 8.25, 8.29, 0.750, 2.50, 0.0800, 7.58, 5.82, 7.57, 2.38, 3.58, 4.98, 1.48, 3.33, 1.32, 5.13 : 9 (**9**)
- 6.33, 2.60, 3.90, 6.80, 3.56, 8.06, 5.75, 8.02, 6.12, 4.04, 8.81, 1.05, 6.90, 3.71, 6.08, 2.25 : 3 (**3**)
- 2.82, 8.69, 5.30, 5.94, 4.80, 2.07, 3.55, 5.57, 4.27, 4.23, 6.44, 2.59, 3.45, 6.74, 7.91, 0.930 : 5 (**5**)

Fig. 3: Example of a test conducted using the trained Memorizer–Recaller network. The input data to the Memorizer are shown in the upper panel. The data numbered 1 through 8 were input into the Memorizer. The input data to the Recaller and output data generated by the Recaller are displayed in the lower panel. These input data were randomly input into the Recaller to obtain the predicted values. The values in parentheses associated with each input and output data in the lower panel represent the correct values.

integers in bold represent the values. The results of this test, in which the output values of the Recaller next to the colon were recorded, are depicted in the lower panel of Figure 3. In this case, the Recaller retrieved the correct values for all keys. This test was repeated 1,024 times, and the mean accuracy was 0.916. This indicates that the Memorizer–Recaller network exhibited good performance even in the tests. Moreover, it could memorize multiple pieces of information in a single memory vector and retrieve relevant information.

Next, we examined the changes in the mean accuracy when the Memorizer–Recaller constructed for $N = 8$ processed different volumes of data. In this experiment, the mean accuracy was computed by repeating the test 1,024 times. The results, listed in Table 3, demonstrate an accuracy exceeding 0.9 for the training dataset comprising 8 data points. How-ever, the performance significantly drops when the amount of processed data increases beyond that. This is a 10-class classification problem; therefore, the accuracy for random answers was 0.1. When the volume of data to be memorized was set to 256, the accuracy was almost the same as that when answering randomly.

| The number of input data | Accuracy |
|---|---|
| 2 | 0.967 |
| 3 | 0.938 |
| 4 | 0.941 |
| 5 | 0.931 |
| 6 | 0.926 |
| 7 | 0.912 |
| 8 | 0.916 |
| 16 | 0.510 |
| 32 | 0.301 |
| 64 | 0.204 |
| 128 | 0.151 |
| 256 | 0.126 |

Table 3: Accuracy of the Memorizer–Recaller, which was trained by setting $N$ to 8 when processing different numbers of input values.

| The order of input data | Accuracy |
|---|---|
| 1 | 0.494 |
| 2 | 0.492 |
| 3 | 0.515 |
| 4 | 0.516 |
| 5 | 0.510 |
| 6 | 0.520 |
| 7 | 0.518 |
| 8 | 0.507 |
| 9 | 0.516 |
| 10 | 0.514 |
| 11 | 0.535 |
| 12 | 0.510 |
| 13 | 0.457 |
| 14 | 0.489 |
| 15 | 0.535 |
| 16 | 0.527 |

Table 4: Accuracy of the Memorizer–Recaller network, which was trained by setting $N$ to 8, with varying orders of input data. The results are based on handling 16 pieces of information. The order of input data is such that 1 represents the oldest input information, and 16 represents the most recent input information.

The Memorizer–Recaller network had distinctive characteristics in the manner of deriving the correct answers. The Recaller generated the correct output for the last 8 pieces of information memorized by the Memorizer; however, it was unable to recall the data provided earlier. For instance, when the input data consisted of 16 items, the Memorizer–Recaller exhibited an accuracy of 0.510, which largely exceeds the accuracy of 0.1, which is the accuracy on random predictions. However, this accuracy did not mean that the Memorizer–Recaller could derive the correct answer with an accuracy of 0.510 uniformly for the data to be memorized. This model derived correct answers for the last 8 data points with an accuracy of approximately 0.9, and the first 8 data points with an accuracy of approximately 0.1, resulting in an average accuracy of 0.510.

Here, we were curious whether this model could remember the most recent 8 pieces of information accurately but completely fail to recall any older information. Therefore, to verify whether there were any patterns in the incorrect predictions, we calculated the accuracy for each piece of input data, from the newest to the oldest, when the input data consisted of 16 items. As shown in Table 4, when the Memorizer–Recaller attempted to memorize 16 pieces of information, the accuracy dropped to around 0.5 regardless of the recency of the information. This indicates that the entire memory was confused.

The Memorizer–Recaller did not complete training when $N = 9$, as shown in Table 2. Additionally, the Memorizer–Recaller, which completed training when $N = 8$, could not retain information remembered 9 pieces ago. It is currently difficult to discuss whether there is a correlation between these two facts, so this will be a future challenge. On the other hand, preliminary analyses show that the value of $N$ at which training is completed decreases or increases correspondingly when the network size is reduced or enlarged. This suggests that the memory performance of the Memorizer–Recaller can be controlled to some extent by the network size. However, we believe that fundamental solutions, rather than such ad hoc measures, are necessary to suppress forgetting in this system. Therefore, further investigation is needed in this regard as well.

Using the proposed learning method, we constructed a system that could memorize multiple pieces of information in a single vector and recall them, after deployment. The Memorizer–Recaller network is similar to an encoder–decoder network, such as a transformer, in terms of the neural network structure, and the computational processing in the Memorizer is an extension of attention neural networks (ANNs) [8] or RNNs. However, by effectively utilizing random data in the learning process and not allowing the same input data to be learned twice, we ensured the AI learned the operation to solve problems instead of learning patterns inherent in the data. Consequently, the network can continue learning after the learning process, which distinguishes it from existing AI. Currently, we succeeded in memorizing 8 pieces of information. Based on the experimental results, the Memorizer–Recaller network trained with $N = 8$ could accurately memorize and recall only the 8 pieces of information it had received so far. When new information was introduced and memorized, it tended to forget the oldest information. This tendency was the same for models trained with smaller $N$ values. For example, the Memorizer–Recaller network trained with $N = 4$ could only remember 4 pieces of information, and the Memorizer–Recaller network trained with $N = 6$ could only remember 6 pieces of information. To improve this, further investigation is needed, not mere ad hoc measures such as increasing the size of the network parameters.

A feature of the Recaller that differentiates it from previous neural network models is that it restores information from a memory vector derived from the input values. Furthermore, the Memorizer differs from the previous neural network models in the ability to append new information to a memory vector, which is a key feature of Memorizer–Recaller networks. A Memorizer–Recaller network possesses Appendable Memory; this memory is a fixed-sized vector which new information can be appended to. We intend to explore the future potential of Appendable Memory vector.

### 3.3 Development of Sorting Algorithm

Next, we explored whether a sorting algorithm can be generated using the Memorizer–Recaller network. As the Memorizer–Recaller network can memorize the input information in a memory vector, we expected that it should generate a sorting algorithm. The rationale behind this hypothesis is as follows: the Memorizer has the capability to sequentially add information to the Appendable Memory, and the Recaller can retrieve information stored in this memory. By designing the training data in a specific manner, as described in the following paragraph, we anticipated that it might be possible to train the Recaller to retrieve the memorized information in ascending order. Consequently, the Memorizer could potentially store the input values in the Appendable Memory while considering their magnitude. As a result, the entire Memorizer–Recaller network could generate a sorting algorithm.

The structure of the network used in the experiment is illustrated in Figure 4. Most structures are the same as those in Figure 2, but the input data of the Recaller differ from key to query. This query is an integer between 0 and $N-1$. Zero is the prompt for the Recaller to output the smallest number among the $N$ input numbers; one is the prompt for the Recaller to output the second smallest number among the $N$ number of inputs; and $N-1$ is the prompt for the Recaller to output the largest number among the $N$ number of inputs. Additionally, the data structure of the key was changed from previous experiments. In previous experiments, the key consisted of a vector of 16 floating-point numbers. However, in this experiment, since we aim to construct a model that solves the problem of sorting real numbers, the key value was set to a single floating-point number. The model sorts these input floating-point numbers. The results of sorting the input values were obtained by arranging the output results. For example, we used the input value of the Memorizer for $N = 5$, as shown in the left panel of Figure 5. The floating-point numbers are the keys to be sorted, and bold characters are the values. The keys are randomly generated from a uniform distribution with a minimum value of 0 and a maximum value of $N$. The Recaller–output sorted sequence for the input data is depicted in the right panel of Figure 5 panel. The queries to the Recaller from top to bottom are 0, 1, 2, 3, and 4.

Learning was conducted in the same manner as described in the subheading 3.2.2. It was terminated when the validation accuracy reached 0.95. By incrementally increasing the value of $N$ from 2, we checked whether learning could be completed normally. We observed that this could be achieved within $500,000$ epochs up to $N = 8$, which was consistent with the results presented in the previous section. Although this value of $N$ can vary marginally depending on the stopping condition of learning, discussing whether the value of $N$ is 8, 7, or 9 is meaningless. However, it can be thought that this extent of memory can only be retained by the Memorizer–Recaller network under the current condition of parameter size and architecture design.

Of the above training on various values of $N$, we conducted an experiment to prove that the $N = 8$ model can correctly sort input. In the experiment, an output was considered correct only when the order of the elements was perfectly sorted. The test was performed $1,024$ times, after which the network could accurately sort in 893 trials. Therefore, the accuracy of the sorting algorithm was 0.872. The generated sorting algorithm cannot provide a perfect answer for all trials and can only sort up to 8 numbers. However, as expected, a sorting algorithm was generated using the Memorizer–Recaller network. There are various types of sorting algorithms, but the worst time complexity of the comparative sorting method, which is the most standard sorting algorithm, is $O(N \log N)$ when the number of inputs is $N$. In contrast, the computational complexity of the sorting algorithm generated using the Memorizer–Recaller is always $O(N)$. This experimental result demonstrates that the Memorizer–Recaller network can store input information in the memory vector and demonstrated its application to an actual problem. Although the generation of the sorting algorithm was successful, as mentioned initially, it is currently unclear whether the Memorizer generates the Appendable Memory by considering the magnitude of the input numbers and whether the Recaller extracts the information in ascending order using that information. Understanding the mechanism by which the Memorizer–Recaller achieves the sorting algorithm may require advancements in the field of AI, particularly in explainable AI.

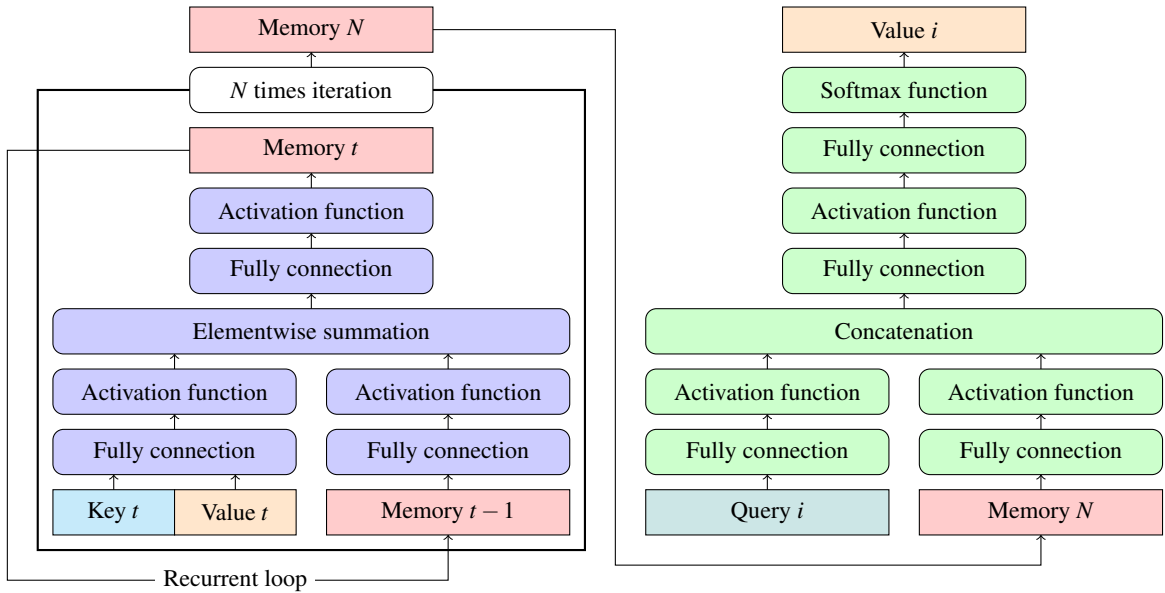

Fig. 4: Illustration of the network when generating a sorting algorithm with the Memorizer–Recaller. The network structure is the same as the previously mentioned Memorizer–Recaller network; however, a part of the input data of the Recaller was changed to a query. In the diagram, rectangles represent data and rounded rectangles represent operations.

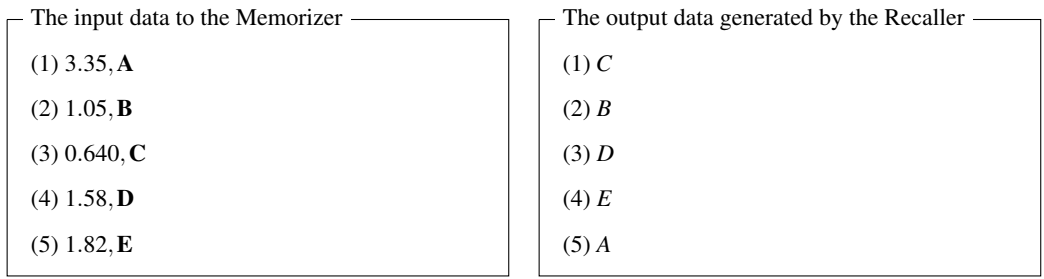

Fig. 5: Example of input and output by a sorting algorithm generated by the Memorizer–Recaller network. The left panel includes the input data to the sorting algorithm. The right panel includes the sorted data by the sorting algorithm.

## 4 Discussion

The aim of this study was to construct a new learning method, which enables AI to learn how to operate, rather than just learning patterns in a training dataset. As an application of the proposed learning method, we developed a system in which the AI can memorize information even after the model has been deployed. Usually, AI acquires knowledge by updating parameter values during machine learning process. The knowledge of AI is essentially embedded in these parameters. On the other hand, the AI that we envision as capable of acquiring knowledge after deployment refers to an AI that can acquire information without changing its parameters after the machine learning process that involves parameter updates. The proposed Memorizer–Recaller network can store input information in the Appendable Memory vector separately and retrieve it using the Recaller. The memory vectors generated by the Memorizer can be saved and utilized subsequently. This memory is a dynamic memory to which new information can be appended. In other words, information can be added to this memory after the machine learning process involving parameter updates, i.e., after deployment. This means that the AI can acquire knowledge after deployment. Furthermore, the Recaller can restore information from these memory vectors. We attempted to construct a sorting algorithm using this network, which also succeeded in demonstrating that the Memorizer could correctly encode the input information into the memory vector and that the Recaller could appropriately utilize that information.

The proposed learning method probabilized the training dataset for each epoch and introduced randomness into the learning process. This prevents a model from learning the inherent patterns of the training dataset. This aspect is a novel idea that can be achieved using random numbers in the learning method.

The Memorizer–Recaller network was developed to allow AI to store information, and we succeeded in memorizing 8 pieces of information. The objective of the study is to examine whether we can develop a system capable of acquiring information even after deployment using the learning method we proposed. Therefore, the number is not a primary concern in the study. However, when actually applying this system for a specific purpose, its limited memory capacity could narrow the scope of its application. As mentioned in subheadings 3.1 and 3.2.2, the Memorizer is a type of RNN. A simple RNN has a forgetting phenomenon where it cannot retain past information. Similarly, the Memorizer experiences some form of forgetting, just like a simple RNN. The initially developed simple RNN [9] has been gradually improved, and Long Short-Term Memory (LSTM) [10] was developed as a type of RNN with reduced forgetting. It may be possible to increase the amount of information that the Memorizer can remember by incorporating mechanisms similar to those introduced in the development of LSTM. For example, as conducted in a study [11], one possible method is to randomly generate the structure of the Memorizer and evaluate their performance to select a structure that reduces the forgetting phenomenon. Further investigation will be necessary to improve the memory capacity of the Appendable Memory system.

The desired operation with the Memorizer can be realized by ANNs. By considering past contextual information, ANNs can receive sequential values as input and output values. In this study, we performed a similar operation using the Memorizer. However, one key distinction between the Memorizer and conventional ANNs is that ANNs cannot append information to a memory vector.

Certain dialogue agents can acquire new knowledge after deployment and have the capability to make conversations based on past interactions stored in their memory. For example, large language models (LLMs) currently deployed typically have this capability. LLMs remember previous questions and their own responses, and act as if they remember past conversations by using a naive method where they concatenate this information with newly input questions [12]. Therefore, one of the major challenges in current LLM development is to increase the input sequence length that LLMs can handle [13]. In contrast, the Appendable Memory system developed in this study memorizes past information into a single vector and recalls the memory from there. As a future application, the Appendable Memory system may be used to improve the performance of dialogue agents as shown in Figure 6. The combination of key and value corresponds to a statement by a particular user, and the memory vector represents the accumulation of these statements stored in the AI system. The key represents some piece of information, and the value is an explanation of this information. The role of a Recaller is to recall the value corresponding to a key based on its memory. If an AI can have this function, it would be able to accumulate knowledge from users or even web searches and grow without changing or updating the parameters of the neural network after deployment.

## 5   Conclusion

This study developed a learning method using randomness of datasets and constructed a system for AI to acquire knowledge even after the model has been deployed. The resulting system is called the Appendable Memory system, consisting of the Memorizer–Recaller network, which generates the Appendable Memory. The proposed learning method probabilized the training dataset for each epoch, thereby prohibiting a model from learning patterns inherent in the training dataset. This differentiates our work from previous learning methods. The advantage of Appendable Memory is that new knowledge is stored in a finite memory vector. This means that the original multiple pieces of information are encoded into a single vector of finite size. If the size of the encoded information is smaller than the original information, it means that compression of information is achieved. Since Appendable Memory is generated by this mechanism, there is no need to retain past information in its original size or format. The Memorizer–Recaller network developed in this study successfully stored separate input information in the Appendable Memory vector and retrieved it using the Recaller.

However, if we apply our system to real-world tasks, the capacity of the memory vector will be a problem, where our system can memorize only up to 8 pieces of information. This poses a significant constraint when applying the technology to real-world scenarios. Therefore, further investigations are required to improve the performance. Despite the preliminary results of this study, the method to train Memorizer–Recaller is fundamentally different from previously

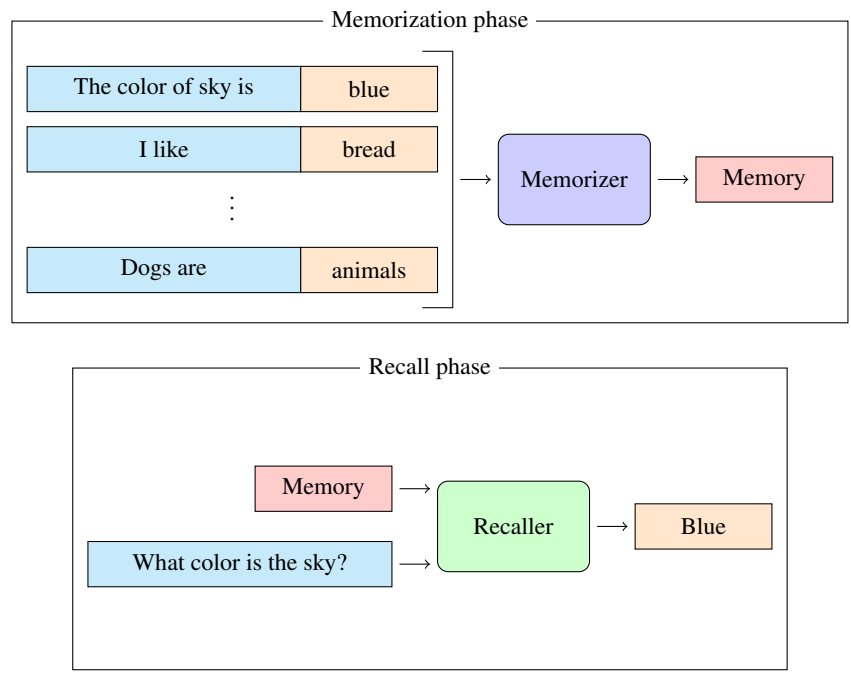

Fig. 6: Example of an application of the Memorizer–Recaller network as a dialogue agent. In the diagram, rectangles represent data, and rounded rectangles represent operations.

implemented neural network learning methods; thus, this approach is considered groundbreaking.

## Funding

This work was supported in part by the Top Global University Project from the Ministry of Education, Culture, Sports, Science, and Technology of Japan (MEXT).